\documentclass[sigconf]{acmart}

\copyrightyear{2026}
\acmYear{2026}
\setcopyright{cc}
\setcctype{by}
\acmConference[KDD '26]{Proceedings of the 32nd ACM SIGKDD Conference on Knowledge Discovery and Data Mining V.2}{August 09--13, 2026}{Jeju Island, Republic of Korea}
\acmBooktitle{Proceedings of the 32nd ACM SIGKDD Conference on Knowledge Discovery and Data Mining V.2 (KDD '26), August 09--13, 2026, Jeju Island, Republic of Korea}
\acmDOI{10.1145/3770855.3818825}
\acmISBN{979-8-4007-2259-2/2026/08}

\AtBeginDocument{%
  }

\usepackage{amsmath}
\usepackage{mathtools}
\usepackage{amsthm}
\usepackage[capitalize,noabbrev]{cleveref}

\usepackage{balance}
\usepackage{natbib}
\usepackage{booktabs} 
\usepackage{enumitem}
\usepackage{balance}
\usepackage{multirow} 
\usepackage{pifont}      
\usepackage{wrapfig}
\usepackage{colortbl}  
\definecolor{gray94}{gray}{.94}
\definecolor{gray90}{gray}{.90}
\newcommand{\grow}[1]{\rowcolor{gray94}{#1}}

\begin{document}

\title{Deciphering Fingerprints of 3D Molecular Surfaces for Accurate Epitope Prediction}


\author{Fang Wu}
\affiliation{\institution{Stanford University}\city{Palo Alto}\country{USA}}

\author{Weihao Xuan}
\affiliation{\institution{The University of Tokyo}\city{Tokyo}\country{Japan}}

\author{Jure Leskovec}
\affiliation{\institution{Stanford University}\city{Palo Alto}\country{USA}}
\authornote{Corresponding author.}

\author{Yejin Choi}
\affiliation{\institution{Stanford University}\city{Palo Alto}\country{USA}}

\author{Li Erran Li}
\affiliation{\institution{Amazon AWS}\city{Palo Alto}\country{USA}}

\renewcommand{\shortauthors}{Fang Wu, Weihao Xuan, Jure Leskovec, Yejin Choi, and Li Li}

\begin{abstract}
    Molecular surfaces encode the geometric and physicochemical patterns that determine antibody-antigen recognition, central to epitope prediction. However, existing methods rely on sequences or backbone structures and struggle to capture discontinuous, surface-driven epitopes. This study presents SurfBind, a surface-centric learning framework for epitope prediction that operates directly on molecular surface representations. SurfBind integrates geometric and physicochemical cues through a Transformer-based architecture with patch-level surface modeling, binder-aware cross-attention, and a hierarchical coarse-to-fine prediction paradigm. Experiments on challenging epitope identification benchmarks, including SAbDab and DB5.5, demonstrate that SurfBind achieves state-of-the-art performance and strong generalization across unseen antibodies and conformational states, highlighting the value of interaction-aware surface modeling for understanding the crucial mechanisms of protein-protein interactions.
\end{abstract}



\maketitle

\section{Introduction}
\begin{figure*}[ht]
    \centering
    \includegraphics[width=0.9\textwidth]{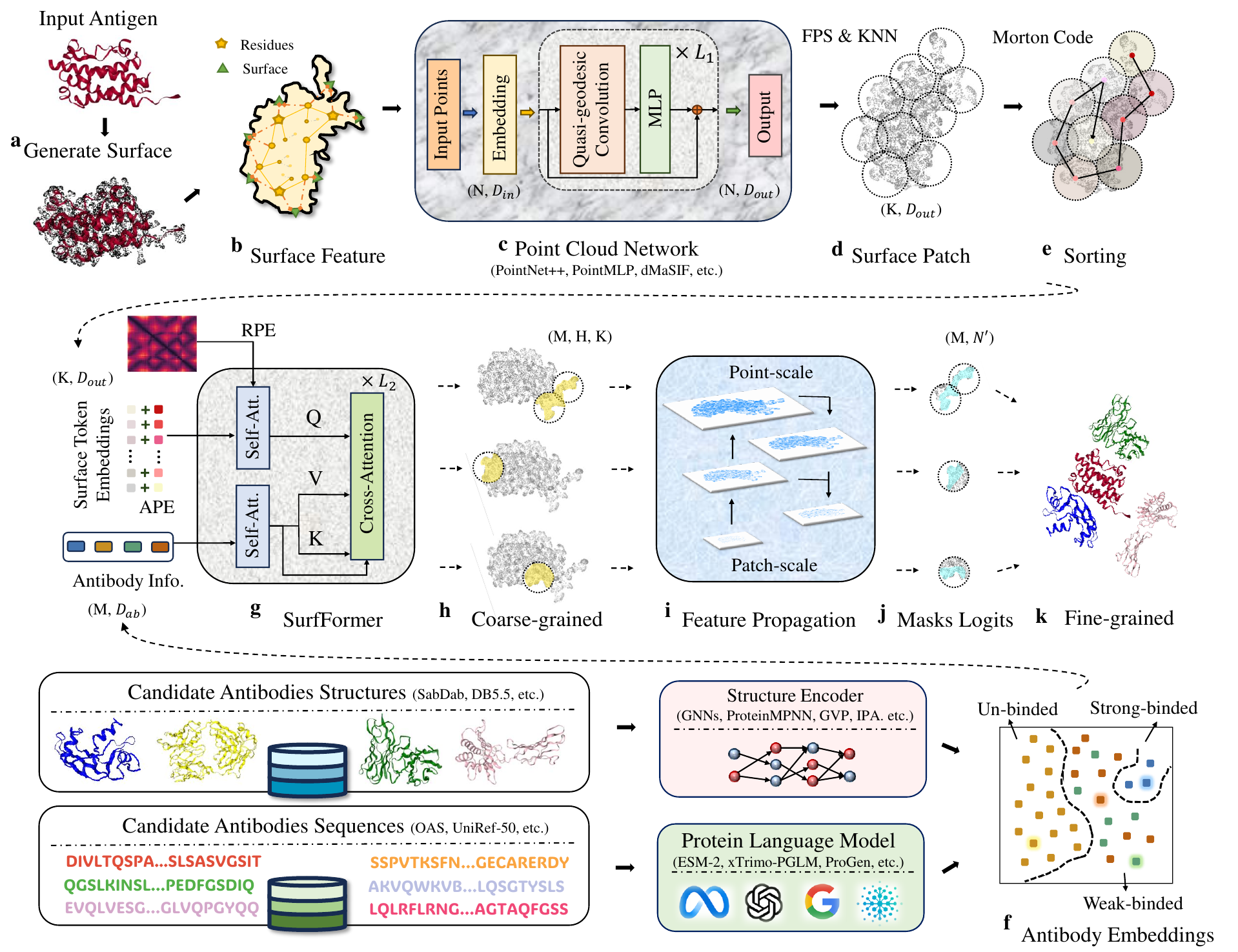}
    \vspace{-1em}
    \caption{\textbf{Schematic overview of our antigen-binding site prediction model.} Firstly, the antigen surface is sampled into a point cloud via a fast sampling mechanism (\textbf{a}), with fine-grained features extracted by a point cloud network (\textbf{b}, \textbf{c}). The point cloud is then downsampled into ordered patches (\textbf{d}, \textbf{e}). In parallel, the antibody is represented using either protein language models (PLMs) or structure encoders (\textbf{f}). The antigen and antibody representations are then fed into SurfFormer to exchange mutual information and achieve binder-awareness (\textbf{g}). Finally, antigen features are propagated from the subsampled patches back to the original points via interpolation (\textbf{i}), enabling multi-resolution epitope prediction at both the point and patch levels (\textbf{h}, \textbf{j}, \textbf{k}).} 
    \label{fig:model_arc}
    \vspace{-1em}
\end{figure*}
Proteins are fundamental components of biological systems, and their most critical functions, particularly in immune recognition and signaling, are often mediated through specific protein–protein interactions (PPIs)~\citep{deng2025predicting,wu2026dynamics}. In antibody-antigen binding, these interactions are governed by epitopes: localized regions on antigen surfaces whose geometric shape and physicochemical composition determine binding specificity and affinity. Accurate epitope prediction is therefore central to antibody engineering, immunotherapy, and vaccine design~\citep{esmaielbeiki2016progress,peters2020t}. However, this task remains challenging due to the complex and heterogeneous nature of epitopes, which are often discontinuous in sequence, sparsely distributed on protein surfaces, and highly sensitive to local surface geometry and chemistry~\citep{zeng2023recent,wu2024semi,wusurfdesign,wu2023molformer,wu2026proteo,wu2022discovering,li2026joint}.

Existing computational methods for epitope prediction primarily rely on sequences or backbone-centric structural features~\citep{rives2021biological,zhang2022protein,clifford2022bepipred,wudiffantiseq,wu20213d,wu2026d,wu2025surface,jiang2025posex}. While effective for capturing global protein properties, these representations often struggle to resolve fine-grained surface patterns that directly mediate antibody–antigen recognition. In contrast, molecular surfaces encode the spatial arrangement and physicochemical complementarity that underlie binding interactions~\citep{mylonas2021deepsurf,riahi2023surface}. Yet surface information is often treated as auxiliary rather than as a first-class modeling target, thereby limiting the ability of existing methods to accurately localize binding sites and generalize to unseen epitopes.

Beyond that, epitope prediction presents additional challenges. First, epitope formation is inherently interaction-dependent: the same antigen surface may expose different binding regions depending on the antibody context, making partner-agnostic predictions unreliable~\citep{potocnakova2016introduction,soria2015overview}. Second, meaningful epitope signals are often subtle and localized, requiring models to reason across multiple spatial scales, from coarse surface regions to fine-grained atomic neighborhoods. Finally, models must generalize across diverse antibodies and antigen families, where binding interfaces can vary significantly in size, shape, and chemical composition~\citep{desai2014t,sanchez2017fundamentals}.

This work introduces SurfBind, which explicitly models binding-relevant surface patterns and cross-molecular dependencies. It bridges the gap between surface pretraining and downstream PPI tasks by integrating geometric surface encoding with binder-aware context modeling and hierarchical coarse-to-fine prediction. Specifically, SurfBind partitions molecular surfaces into irregular local patches that respect the sparsity and redundancy of surface point clouds and organizes them using Morton ordering to enable efficient global reasoning. SurfFormer++ is then employed to model long-range dependencies among surface patches and to incorporate geometric priors. Crucially, SurfBind extends beyond single-surface encoding by introducing binder-aware cross-attention, enabling the explicit exchange of information between interacting molecular partners. To encourage interaction-aligned representations, SurfBind leverages discrete latent modeling and multi-level reconstruction objectives that target not only point statistics but also surface geometry and physicochemical properties.

Evaluation on standard epitope prediction benchmarks demonstrates that explicitly modeling surface-binder interactions yields improved accuracy, stronger generalization to unseen epitopes, and greater robustness across diverse antibody contexts. These results highlight the importance of interaction-driven surface modeling for epitope discovery and advance the state of the art in computational antibody-antigen interface recognition. 

\section{Method}
\subsection{Preliminaries and Mathematical Notations}

\paragraph{Task Description} 
Epitopes, known as antigenic determinants (ADs), are specific regions on antigens' surfaces, which activate the human immune system against pathogens or abnormal cells~\citep{zeng2023recent,wu2026semi}. Their characterization and identification are significant for designing therapeutic or diagnostic antibodies, developing immunodiagnostic tests, and advancing epitope-based peptide vaccines to combat infectious diseases~\citep{bukhari2022machine}. Additionally, ADs may influence, yet are often overlooked, in improving the efficacy of RNA vaccines. Their properties determine whether RNA vaccines can elicit an immune response and which responses will ensue.  

Epitopes are classified into two categories: B-cell and T-cell epitopes. B-cell epitopes (BCEs) are antigen fragments recognized by B cells and feature solvent-exposed regions and can be classified as conformational or linear. Linear BCEs consist of consecutive peptides and residues, while conformational BCEs comprise patches of solvent-exposed atoms from non-sequential residues, termed continuous and discontinuous BCEs, respectively. Experimental techniques, such as peptide microarrays and phage display libraries, help identify linear BCEs~\citep{qi2021antibody}. 
However, approximately 90\% of native BCEs are discontinuous, and mapping conformational BCEs without a complex structure is more difficult as their component residues may be far apart in sequences but spatially co-located within protein structures. Hydrogen/deuterium exchange experiments can infer this sort of BCEs but are confounded by allosteric structural perturbation when the binding effect extends beyond the binding site~\citep{deng2017suppressing}. Alternatively, computational methods like homology modeling, docking simulations, and molecular dynamics simulations are employed. Despite their success, most conventional approaches are time-consuming and require expertise in protein structure and function. 

\paragraph{Protein surface representation.}
We follow established surface construction and preprocessing pipelines~\citep{sverrisson2021fast,mylonas2021deepsurf,stebliankin2023evaluating,li2023geobind,wu2024surface} to achieve effective protein surface learning. A protein with $N$ atoms is represented as $\mathcal{V}^a = \{(\mathbf{x}^a_i, \mathbf{t}^a_i)\}_{i=1}^{N}$, where $\mathbf{x}^a_i \in \mathbb{R}^3$ denotes atom coordinates and $\mathbf{t}^a_i\in\mathbb{R}^6$ encodes their one-hot chemical types in the list $[\mathrm{C}, \mathrm{H}, \mathrm{O}, \mathrm{N}, \mathrm{S}, \mathrm{Se}]$. Protein surfaces are modeled as level sets of a smooth signed distance function (SDF) defined over atom centers.

Each surface point $\mathbf{x}^s_i \in \mathbb{R}^3$ is initialized by stochastic sampling around atom coordinates and projected onto a target SDF level set via gradient-based optimization. Surface normals $\mathbf{n}^s_i$ are computed as normalized SDF gradients at $\mathbf{x}^s_i$. After removing interior points, the resulting protein surface is represented as an oriented point cloud $S = \{(\mathbf{x}^s_i, \mathbf{n}^s_i)\}_{i=1}^{M}.$ Every surface point is augmented with a chemical feature vector $\mathbf{h}^s_i \in \mathbb{R}^{\phi_h}$. To compute $\mathbf{h}^s_i$, residue-level information is aggregated from $K_{\mathrm{res}}$-nearest residues$\{(\mathbf{x}^R_j, \mathbf{t}^R_j)\}_{j=1}^{K_{\mathrm{res}}}$ based on $C_\alpha$ distances, using a lightweight geometric aggregation network. This residue-centric representation offers an efficient approximation of local chemical environments while maintaining strong empirical performance.

\paragraph{Surface patch partition and ordering.} To enable scalable modeling, the surface point cloud
$\mathbf{X}^s = \{\mathbf{x}^s_i\}_{i=1}^{M}$ is partitioned into local patches. Specifically, a subset of patch centers $\mathbf{X}^{\mathrm{c}} \in \mathbb{R}^{\rho M \times 3}$ is selected using farthest point sampling (FPS) with downsampling ratio $\rho$. For each center point, a local patch $\mathbf{X}^{\mathrm{p}} \in \mathbb{R}^{\rho M \times K_{\mathrm{p}} \times 3}$ is formed by selecting its $K_{\mathrm{p}}$ nearest neighbors from $\mathbf{X}^s$.

As point clouds lack a canonical ordering, we impose a geometric sequence structure over patches. Patch centers $\mathbf{X}^{\mathrm{c}}$ are mapped to a 1D ordering using a Morton (Z-order) space-filling curve~\citep{morton1966computer}, producing an index sequence $\mathcal{O}\in \mathbb{N}^{\rho M \times 1}$. Patches $\mathbf{X}^{\mathrm{p}}$ are arranged according to $\mathcal{O}$, which preserves local spatial coherence while enabling sequence-based processing in downstream models~\citep{chen2023pointgpt}.

\subsection{Backbone Architecture}
To capture the hierarchical granularity inherent in protein surfaces (\emph{i.e.}, point-level and patch-level features), we hierarchically extract surface details at multiple scales. We then propose SurfFormer++, which incorporates a cross-attention module to enable sufficient information exchange between ligand and receptor patches.

\paragraph{Point cloud network.} We employ a standard surface point cloud network to extract local, point-wise surface representations from the oriented surface point cloud $S = \{(\mathbf{x}^s_i, \mathbf{n}^s_i)\}_{i=1}^{M}$. The network follows the quasi-geodesic convolution paradigm~\citep{sverrisson2021fast}, where each surface point $\mathbf{x}^s_i$ is equipped with a local orthonormal frame $(\mathbf{n}^s_i, \mathbf{u}^s_i, \mathbf{o}^s_i)$~\citep{duff2017building} and aggregates features from a geodesic neighborhood $\mathcal{N}(i)$, which is determined by the filter window size $\sigma_\mathrm{d}$. Neighbor interactions, parameterized using relative geometric descriptors $\mathbf{p}_{ij}=  \left(\mathbf{x}^s_i-\mathbf{x}^s_j\right)^\top \cdot \left[{\mathbf{n}}^s_i\oplus {\mathbf{u}}^s_i\oplus {\mathbf{o}}^s_i\right]$, are defined in the local coordinate system and weighted by a Gaussian function $w(\mathrm{d}_{ij})$ of an approximate geodesic distance $\mathrm{d}_{ij}$. Stacking $L_1$-layer operator yields point-level surface features $\{\mathbf{h}^s_i\}_{i=1}^{M}$, subsequently used for patch-level modeling.

\paragraph{SurfFormer++.} Patch-level representations are constructed by aggregating point-wise features within each surface patch. Specifically, for patch $\mathbf{X}^{\mathrm{p}}_i$, point features $\{{\mathbf{h}^s_j}^{(L_1)}\}$ are pooled and mapped to an initial patch embedding $\mathbf{h}^{\mathrm{p}}_i \in \mathbb{R}^{\phi_\mathrm{p}}$. Global interactions between patches are modeled using a modified Transformer~\citep{vaswani2017attention}, referred to as SurfFormer, comprising $L_2$ layers.

Each SurfFormer layer applies multi-head self-attention over patch features $\{\mathbf{h}^{\mathrm{p}}_i\}$, augmented with geometric structural embeddings derived from approximate geodesic distances $\mathrm{d}_{ij}$ via radial basis functions~\citep{schutt2018schnet}. These embeddings serve as invariant relative positional encodings, enabling geometry-aware attention while preserving rotational and translational invariance. In addition, the geometric ordering $\mathcal{O}$ is used to assign absolute sinusoidal positional embeddings to patches, which are then added at each layer to facilitate modeling of global context.

\paragraph{Binder-aware Cross-attention Block} Incorporating binder information holds paramount importance in real-world applications. For instance, a more comprehensive epitope prediction task involves identifying the antigen's interacting residues given a particular antibody~\citep{chen2024tepcam}. To this end, cross-attention is a widely used technique that facilitates the mutual exchange of features between two components. 

Notably, binders may only possess 1D sequences without crystal structures. In such cases, protein language models (PLMs) like ESM-2~\citep{lin2022language} enable the acquisition of sequence-level representations denoted as $\mathbf{h}^{\mathrm{PLM}}_{\mathrm{lig}}\in \mathbb{R}^{\phi_{\mathrm{PLM}}}$. Then, an MLP is appended to increase the channel dimension, and the output representation is reshaped to a suitable number of patch vectors. In the ideal scenario where both ligand and receptor structures are accessible, their patch features are written as $\left\{\mathbf{h}^\mathrm{p}_{j,\mathrm{lig}}\right\}_{j=1}^{\rho M'}$ and $\left\{\mathbf{h}^\mathrm{p}_{i,\mathrm{rec}}\right\}_{i=1}^{\rho M}$, respectively, where we suppose $M'$ patches in the ligand surface cloud $\mathbf{X}^s_{\mathrm{lig}}$. The attention score is then computed as
\begin{equation}
    e_{ij}^{(l)}= \frac{\left({\mathbf{h}^\mathrm{p}_{i,\mathrm{rec}}}^{(l)}\mathbf{W}_Q\right)\left({\mathbf{h}^\mathrm{p}_{j,\mathrm{lig}}}^{(l)}\mathbf{W}_K\right)^\top}{\sqrt{\phi_\mathrm{p}}}.    
\end{equation}
Here, we omit the geometric structural embedding term $ \mathbf{r}_{ij}$ since the relative distances between ligands and receptors are typically unknown. This module quantifies the influence of a ligand patch $\mathbf{h}^\mathrm{p}_{j,\mathrm{lig}}$ on a receptor patch $\mathbf{h}^\mathrm{p}_{i,\mathrm{rec}}$ by generating attention scores $e_{ij}$ that help identify relevant interactive patch pairs within a complex. Further details on SurfFormer++ are provided in App.~\ref{app:surfformer}.

\subsection{Pretraining on Molecular Surfaces}
By relying less on annotation, self-supervised learning has significantly advanced domains such as language, vision, and life sciences. Recently, Surface-VQMAE~\citep{wu2024surface} performs masked autoencoder (MAE)~\citep{he2022masked} on molecular surfaces, randomly masking a portion of surface patches and using an auto-encoder to reconstruct surface features. We extend their framework by introducing carefully designed recovery targets. 

\paragraph{Masking and tokenization.} Surface patches are masked independently to account for patch overlap, with a masking ratio $\delta$. Masked and visible patch coordinate sets are denoted as $\mathbf{X}^{\mathrm{p,m}}$ and $\mathbf{X}^{\mathrm{p,vis}}$, respectively. In line with empirical practice, relatively high masking ratios ($\delta \geq 50\%$) are used without degrading performance. Instead of a shared mask embedding, masked patch tokens are replaced by latent code embeddings using a vector-quantized (VQ) formulation. We use the discrete VAE paradigm~\citep{van2017neural,ramesh2021zero} to establish a codebook $\mathcal{Q}=\{\mathbf{e}(i)\}_{i=1}^{N_B}$, which contains a group of embeddings $\mathbf{e}(i)\in\mathbb{R}^{\phi_{\mathrm{p}}}$ and sample latent patch representations $\mathbf{z}^{\mathrm{p,m}}_i$ via a Gumbel-Softmax relaxation~\citep{jang2016categorical}. This relaxed posterior allows uncertainty to be expressed over masked tokens while remaining fully differentiable. When $N_B=1$, the formulation reduces to the standard MAE setup.

\paragraph{Geometric Decoding Targets.} Visible patch embeddings and sampled latent codes are merged as ${\mathbf{H}^{\mathrm{p}}}^{(0)}=\mathbf{H}^{\mathrm{p,vis}}\oplus\mathbf{Z}^{\mathrm{p,m}}$ and processed by SurfFormer and lightweight decoders, producing final patch representations ${\mathbf{H}^{\mathrm{p}}}^{(L_2)}$. Prediction targets are defined over masked patches and include point-level statistics and surface geometry. In particular, masked patch coordinates are reconstructed using a simple MLP head~\citep{li2023masked,chen2023pointgpt}, and surface curvature descriptors are predicted based on local covariance analysis~\citep{mitra2003estimating,tian2023geomae}. These geometric targets are invariant to rigid transformations.

\paragraph{Physichemical Decoding Targets} In addition to geometry, we propose another group of chemical features~\citep{leem2022deciphering} as the pretraining targets denoted as $\boldsymbol{\varrho}$, including \textbf{hydrogen bond acceptor potential and proton donors} and \textbf{hydropathy}. Specifically, the locations of free electrons and potential hydrogen-bond donors on the molecular surface were computed using a hydrogen-bond potential as a reference. Vertices on the molecular surface whose closest atom is a polar hydrogen, nitrogen, or oxygen were considered potential donors or acceptors in hydrogen bonds. Then, a value from a Gaussian distribution was assigned to each vertex depending on the orientation between the heavy atoms. These values range from $-1$, which represents the optimal position for a hydrogen bond acceptor, to $+1$, which represents the optimal position for a hydrogen bond donor. At the same time, each vertex was assigned a hydropathy scalar value according to the Kyte and Doolittle scale of the amino acid identity of the atom closest to the vertex. These values, in the original scale, ranged from $-4.5$ (hydrophilic) to $+4.5$ (most hydrophobic) and were then normalized to $[-1,1]$. Notably, we employ three lightweight MLPs to decode these physicochemical targets, thereby forcing the encoder to embed more semantic information in the surface point clouds. 

\paragraph{Training Losses.} The overall training objective consists of four parts: the typical losses to recover the coordinates, curvatures, and chemical features for each surface patch, as well as the Kullback-Leibler (KL) divergence to approximate the desired latent distribution $p(.)$. Rigorously, the total loss $\mathcal{L}$ is $\nu_1 \mathcal{L}_{\mathrm{rec}}\left(\mathbf{X}^{\mathrm{p,m}}, \hat{\mathbf{X}}\right) + \\ \nu_2 \mathcal{L}_{\mathrm{cur}}\left(\boldsymbol{\psi}, \hat{\boldsymbol{\psi}}\right) + \nu_3 \mathcal{L}_{\mathrm{chem}}\left(\boldsymbol{\varrho}, \hat{\boldsymbol{\varrho}}\right) + \nu_4 \mathcal{L}_{\mathrm{KL}}\left(q\left(\mathbf{Z}^{\mathrm{p,m}}|\mathbf{H}^{\mathrm{p,m}}\right), p(\mathbf{Z}^{\mathrm{p,m}})\right)$, where $\{\nu_i\}_{i=1}^4$ are pre-defined hyperparameters to balance the weights of different loss terms. $p(.)$ is the prior on the latent space and is usually initialized to a uniform distribution over all codebook vectors. $\mathcal{L}_{\mathrm{cur}}(.)$ and $\mathcal{L}_{\mathrm{chem}}(.)$ are both supervised via a root mean squared error (RMSE). Meanwhile, the reconstruction loss $\mathcal{L}_{\mathrm{rec}}(.)$ is formulated using the $l_2$-norm Chamfer distance~\citep{fan2017point} as $\frac{1}{\delta\rho M K_\textrm{p}} \sum_{i=1}^{\delta\rho M} \bigg (\\\sum_{a\in {\hat{\mathbf{X}}}_i} \min_{b \in \mathbf{X}^{\mathrm{p,m}}_i}\|a - b\|^2_2  + \sum_{b \in \mathbf{X}^{\mathrm{p,m}}_i} \min_{a\in {\hat{\mathbf{X}}}_i}\|a - b\|^2_2\bigg)$.

\subsection{Coarse-to-fine Interface Prediction}
\paragraph{Point Feature Propagation.} In the patch partition module, the original point set undergoes subsampling. However, in epitope discovery, it is necessary to identify all original surface points. Therefore, we upsample surface patches to gradually restore the fine-grained representations of the complete surface point cloud. 

To this end, we employ a technique inspired by PointNet++~\citep{qi2017pointnet++}, wherein features are propagated from subsampled patches to the initial points. It is realized by interpolating feature values of $\rho M$ surface patches at the coordinates of the $M$ surface points. Here, we adopt the inverse geodesic distance weighted average based on $K_\textrm{in}=3$ nearest neighbors for this interpolation and attain:
\begin{equation}
    \mathbf{h}_i^{s,\textrm{in}} = \frac{\sum_{j=1}^{K_\textrm{in}} \mathrm{d}_{ij}^{-2} {\mathbf{h}_j^\mathrm{p}}^{(L_2)}}{\sum_{j=1}^{K_\textrm{in}}\mathrm{d}_{ij}^{-2}} , i=1,..., M.
\end{equation}
The interpolated features on $M$ points are then concatenated with skip-linked features from the point cloud network as $\mathbf{h}_i^{s,\textrm{in}} \oplus {\mathbf{h}^s_i}^{(L_1)}\in \mathbb{R}^{2\phi_p}$. A few shared fully-connected and ReLU layers are applied to update each surface point's feature vector, resulting in ${\mathbf{h}_i^{s}}'$. The total loss $\mathcal{L}_{\mathrm{ep}}$ is a weighted sum of the coarse-scale (\emph{i.e.}, patch-level) $\mathcal{L}_{\mathrm{p}}$ and the fine-scale (\emph{i.e.}, point-level) $\mathcal{L}_{\mathrm{s}}$ with a balance term $\zeta$ as: 
\begin{equation}
\begin{split}
    &\mathcal{L}_{\mathrm{ep}} = \mathcal{L}_{\mathrm{s}} + \zeta \mathcal{L}_{\mathrm{p}} \\ &= \mathrm{BCELoss} \left( \mathbf{Y}^{\mathrm{s}}, \mathrm{MLP}\left({{\mathbf{H}^{s}}'}\right)\right) + \zeta \mathrm{BCELoss} \left(\mathbf{Y}^{\mathrm{p}}, \mathrm{MLP}\left({\mathbf{H}^\mathrm{p}}^{(L_2)}\right) \right), 
\end{split}
\end{equation}
where $\mathbf{Y}^{\mathrm{p}}\in \mathbb{R}^{\rho M}$ and $\mathbf{Y}^{\mathrm{s}}\in \mathbb{R}^{M}$ are the coarse- and fine-grained ground truth epitope labels, respectively. A binary cross-entropy loss function (BCELoss) is utilized for supervision. Remarkably, we adopt soft labels for $\mathbf{Y}^{\mathrm{p}}$ to indicate the degree or likelihood of surface patches to be the epitope, which is computed as the ratio of epitope points in each surface patch as $\mathbf{y}^{\mathrm{p}}_i = 1/K_\mathrm{p} \sum_{\mathbf{x}_j^s\in \mathbf{x}^\mathrm{p}_i} \mathbf{y}^s_j$.  
Due to the class imbalance issue (\emph{i.e.}, the number of non-epitopes is far more than the number of epitopes), we investigate the focal loss~\citep{lin2017focal} to focus learning on hard misclassified examples, but observe no significant improvements. 

\paragraph{Extension to Multi-scale Surface Representations}  Our approach can be extended to $S$ scales ($S>2$) for hierarchical learning by a successive surface patch partition strategy. Specifically, we regard $\mathbf{X}^s$ as the 0-scale. Then, for the $i$-th scale ($1\leq i\leq S$), the center point $\mathbf{X}^{\mathrm{c}_i} \in \mathbb{R}^{\Pi_{j=1}^i{\rho}_j M \times 3}$ can be produced by repeatedly using the FPS with a downsampling ratio of $\rho_i$ (typically $\rho_i\leq \rho_{i+1}$), and the corresponding patch $\mathbf{X}^{\mathrm{p}_i} \in \mathbb{R}^{\Pi_{j=1}^i{\rho}_j M \times K_\textrm{p} \times 3}$ is obtained by aggregating the neighboring $K_\textrm{p}$ points. Consequently, by recursively back-projecting, the masked and visible patches of all scales are acquired, denoted as $\{\mathbf{X}^{\mathrm{p}_i,\mathrm{m}}, \mathbf{X}^{\mathrm{p}_i,\mathrm{vis}} \}_{i=1}^S$. And the coarse-scale loss consists of multiple items as $\sum_{i=1}^S  \mathrm{BCELoss} \left(\mathbf{Y}^{\mathrm{p}_i}, \mathrm{MLP}\left({\mathbf{H}^{\mathrm{p}_i}}^{(L_2)}\right) \right) $.

\section{Results} 

\subsection{Binding Site Prediction }

\paragraph{Data Preprocessing}
The pretraining data for SurfBind were derived from \textbf{PDB-REDO}~\citep{joosten2014pdb_redo}. The \textbf{SAbDab} database~\citep{dunbar2014sabdab}, as of 23 September 2023, was used to evaluate BCE predictions. X-ray crystal structures of Ab-ag complexes binding to proteins with a resolution of 3.0\textup{\AA} or better were filtered. We deleted samples that lacked antigens or had incomplete antibodies, including those missing a heavy or light chain. We also removed additional illegal data points with antigen chain lengths <10, yielding 5,531 complex structures. The remaining subset was clustered into 658 groups based on the antigen sequence identity of 30\% using MMseqs2~\citep{steinegger2017mmseqs2}. These clusters were then split into training, validation, and test sets, with approximately 80\%, 10\%, and 10\% allocated by sequence identity, antigen species, and binder count, respectively. To be specific, most antigens in SAbDab have a single unique antibody, whereas antigens such as the HIV gp120 glycoprotein and the SARS-CoV-2 spike protein have many known binders whose sequences are highly dissimilar. To assess the specificity of Ab-ag binding sites, clusters of single-binder antigens were randomly selected as validation and test samples based on the super-clusters of antigen sequences, with the remaining samples used for training. Meanwhile, the super-clusters of those multi-binder antigens were randomly split into training, validation, and test sets in proportions of 40\%, 30\%, and 30\%, respectively. Ultimately, we had 4,572, 548, and 411 samples under this split. For label computation, an amino acid from an antigen is labeled as BCE if at least one heavy atom is within 4\textup{\AA} of another heavy atom from the antibody within the biological assembly~\citep{tubiana2022scannet}. 

\paragraph{Baselines and Evaluation Metrics}
We compared our mechanism with prior studies categorized into three types: sequence-based, structure-based, and surface-based. Remarkably, major BCE prediction methods including \textbf{CBTOPE}~\citep{ansari2010identification}, \textbf{BepiPred-2.0}~\citep{jespersen2017bepipred}, \textbf{BepiPred-3.0}~\citep{clifford2022bepipred}, \textbf{Seppa-3.0}~\citep{zhou2019seppa}, \textbf{Epitope3D}~\citep{da2022epitope3d}, \textbf{ScanNet}~\citep{tubiana2022scannet}, \textbf{PeSTo}~\citep{krapp2023pesto}, \textbf{SEMA-2.0}~\citep{shashkova2022sema}, \textbf{DiscoTope-3.0}~\citep{hoie2024discotope}, \textbf{MaSIF}~\citep{gainza2020deciphering}, and \textbf{dMaSIF}~\citep{sverrisson2021fast} were designed and trained in a partner-agnostic way and forecasted unified binding sites, namely, all potential epitopes. In contrast, approaches like \textbf{AF-Multimer}~\citep{evans2021protein},  \textbf{Pair-EGRET}~\citep{alam2023pair}, \textbf{WALLE}~\citep{liu2024asep}, and \textbf{SEPPA-mAb}~\citep{qiu2023seppa} allowed for antibody-specific BCE discovery. 
Notably, outputs of surface-based algorithms such as SurfBind were represented as meshes or point clouds, whereas others produced residue-level predictions. To ensure a fair comparison across different representation scales, we mapped all surface-based predictions to the residue level using a simple nearest-neighbor rule. Specifically, the prediction score for any residue was set equal to the score of its closest surface point, patch, or mesh face.
Additionally, following~\citet{da2022epitope3d,cia2023critical}, we disregarded buried residues and only considered residues close to the surface based on the relative solvent accessible surface area (RSA) threshold of 15\%. This optimization focuses the BCE task on the most relevant residues for antibody binding. More details are in App.~\ref{app:exp}. 
\begin{table}[t]
\caption{Performance of various algorithms for the BCE discovery, where AB-S is the abbreviation of antibody-specific. }  
\label{tab:ep_sabdab} 
\centering
\resizebox{0.9\columnwidth}{!}{
\begin{tabular}{l|c|ccccc} \toprule
    Method & AB-S & AUC-ROC & AUC-PR & BAcc & F1 \\ \midrule
    \multicolumn{5}{l}{\textbf{Sequence-based}} \\
    CBTOPE & \ding{55}  & 45.71 & 9.88 & 46.90 & 12.30 \\
    BepiPred-2.0 & \ding{55} & 52.69 & 10.74 & 50.81 & 14.55 \\
    BepiPred-3.0 & \ding{55} & 58.17 & 10.98 & 52.66 & 15.04 \\
    Seppa-3.0 & \ding{55} & 66.73 & 14.53 & 64.52 & 28.80 \\
    AF-Multimer & \ding{51} & 69.80 & 17.43 & 67.08  & 33.45 \\
    \multicolumn{5}{l}{\textbf{Structure-based}} \\
    Epitope3D & \ding{55} & 43.85 & 9.56 & 42.09 & 11.84 \\
    ScanNet & \ding{55} & 70.38 & 17.84 &  68.61 & 34.02 \\
    PeSTo & \ding{55} & 72.45 & 19.85 & 71.09 & 36.68 \\ 
    SEMA-2.0 & \ding{55} &  73.69 &  17.40 & 72.58 & 38.09\\
    DiscoTope-3.0 & \ding{55} & \underline{75.18} &  \underline{20.07} & \underline{75.99} & \underline{38.57} \\
    Pair-EGRET &  \ding{51} & 71.09 & 18.32 & 71.83 & 35.45 \\ 
    WALLE & \ding{51} & 70.83 & 18.05 & 69.46 & 34.17 \\
    \multicolumn{5}{l}{\textbf{Surface-based}} \\
    MaSIF & \ding{55} & 61.25 & 12.07 & 64.11 & 18.32 \\
    dMaSIF & \ding{55} & 64.13 & 13.60 & 66.18 & 20.74 \\ 
    SEPPA-mAb &  \ding{51} & 72.38 & 19.04 & 74.63 & 37.02 \\ \midrule 
    \grow{SurfBind} & \ding{51} & $\mathbf{81.62}$ & $\mathbf{30.57}$ & $\mathbf{76.98}$ & $\mathbf{42.95}$ \\ \bottomrule
\end{tabular}}    
\vspace{-1em}
\end{table}
\begin{figure*}[t]
    \centering
    \includegraphics[width=0.8\textwidth]{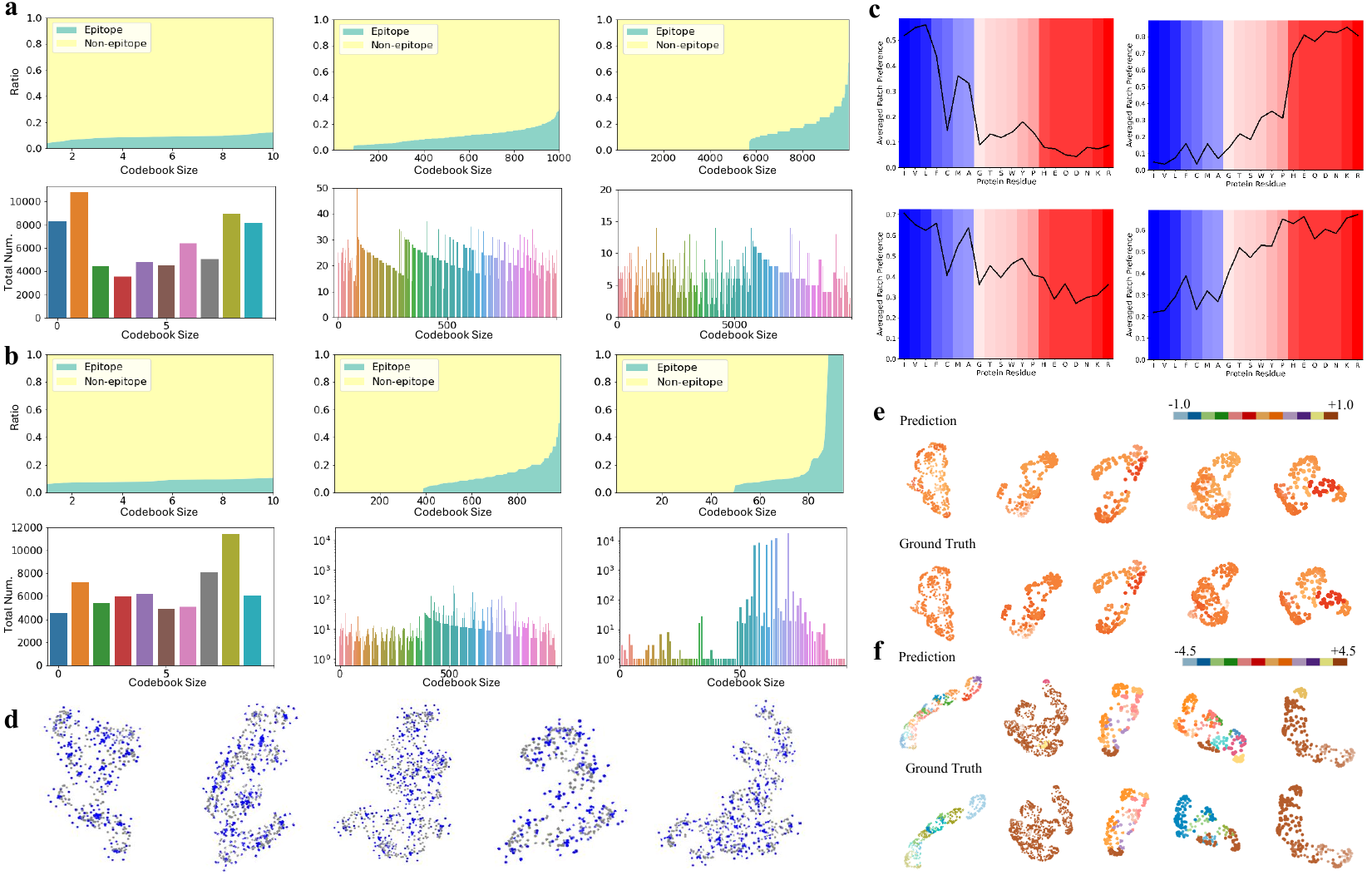}
    \vspace{-1em}
    \caption{\textbf{(a, b)} BCE vs. non-BCE ratio distributions based on unsupervised SurfBind representations using different codebook sizes, ranging from 10, 1000, to 10000, ordered by descending epitope ratio. Bar plots display the number of patches assigned to each cluster. Fig. \textbf{a} used only point coordinate reconstruction as the pretext task, whereas Fig. \textbf{b} added the surface geometry and chemical property prediction tasks for pretraining. \textbf{(c)} Average residue type preferences of representative patch clusters over forty multi-binder antigens. Background color ranges from blue (hydrophobic residues) to red (hydrophilic residues) based on Kyte-Doolittle hydropathy scales. \textbf{(d)} Examples of the point coordinate reconstruction pretext task. The grey and blue ones are the ground truth and predicted surface point clouds, respectively. \textbf{(e)} Examples of the hydrogen bond donor/acceptor prediction pretext task, where colors indicate values from -1 to 1. \textbf{(f)} Examples of the hydrophobicity prediction pretext task, where colors show values from -4.5 (hydrophilic) to 4.5 (hydrophobic). } 
    \label{fig:unsper}
\end{figure*}

\begin{figure*}[t]
    \centering
    \includegraphics[width=\textwidth]{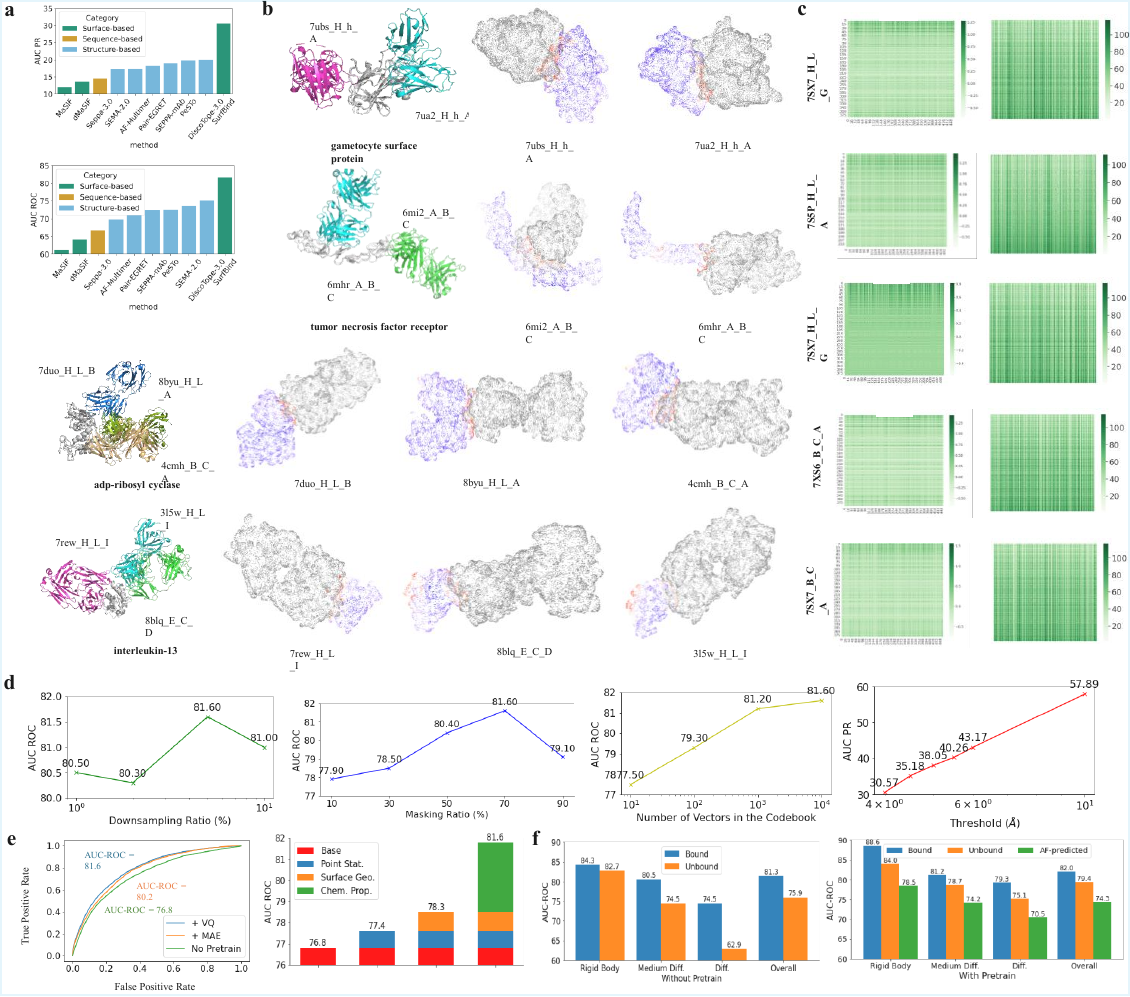}
    \vspace{-2em}
    \caption{\textbf{(a)} Performance comparison for antibody-specific BCE prediction on Sabdab. \textbf{b}. Case study visualization of four protein complexes. Different antibodies can bind to entirely different regions of the same antigen (grey structures). Predicted epitopes (red) and non-epitopes (purple) are shown on the antigen point clouds. \textbf{c}. Explainability of SurfBind attention. Attention score maps were highly negatively correlated with distance maps between antigen residues (rows) and antibody residues (columns), where antigen residues were sorted by their minimum distance to the antibody. \textbf{d}. Hyperparameter search. We reported fine-tuned AUC-ROCs across various masking ratios, downsampling ratios, and codebook sizes, and AUC-PRs across different binding-site distance thresholds. \textbf{e}. Ablation studies. We documented the performance gains achieved by different pretraining techniques and auxiliary construction targets.  \textbf{f}. Influence of conformational changes on SurfBind. Bar plots showed results on bound, unbound, and predicted antigen structures from DB5.5. } 
    \label{fig:super}
\end{figure*}

\paragraph{Quantitative Comparison for BCE Discovery} Tab.~\ref{tab:ep_sabdab} and Fig.~\ref{fig:super} \textbf{a} document the main results. SurfBind achieves the best overall performance on the primary metrics, with an AUC-PR of 0.305 and an F1 score of 0.429, which are considered primary because they balance precision and recall. This represented improvements of 75.38\%, 66.86\%, and 60.55\% over antibody-specific methods using AF-Multimer, Pair-EGRET, and SEPPA-mAb, respectively. Remarkably, the AUC-PRs of the classic surface-based models MaSIF and dMaSIF were only 0.120 and 0.136, respectively, far worse than SurfBind. 
Additionally, AF-Multimer, while highly effective for structure prediction, is not optimized for epitope localization and performs poorly under this evaluation protocol, with an AUC-ROC of 0.698. This aligns with the recent study~\citep {polonsky2023evaluation}, which concluded that AF-Multimer was limited in its ability to predict Ab-ag complexes and to map epitopes. Potential reasons for this failure included the lack of paired MSAs, low sensitivity to antibody sequences, and unimodal rather than multimodal prediction~\citep{tubiana2022scannet}. 
In general, sequence-based algorithms (average AUC-ROC = 0.586) were inferior to structure-based models (average AUC-ROC = 0.677), highlighting the importance of 3D complementary information for the BCE prediction. It is also noteworthy that some of the latest structure-based methods, such as DiscoTope-3.0, have incorporated PLM embeddings~\citep {evans2021protein} and have observed significant improvements. Integrating knowledge from PLMs is expected to significantly enhance SurfBind~\citep{wu2023integration}. 

\paragraph{Binder-awareness of SurfBind}
In addition to the overall quantitative metrics, we investigated SurfFormer's ability to distinguish different antibody binding sites on the same antigen. We selected four representative antigens from the SAbDab-test set, each with multiple partner antibodies that bind to significantly different regions (see Fig.~\ref{fig:super} \textbf{b}). The gametocyte surface protein (GSP) is important for male/female gamete fusion and exflagellation, and it interacts with host erythrocytes. Two antibodies in PDB \textsl{7ubs} and \textsl{7ua2} targeting GSP had different lengths of complementarity-determining regions (CDRs) in the variable fragment (Fv), with 55.8\% sequence identity. They are bound to opposite sides of GSP, and SurfBind successfully located their distinct binding sites. Tumor necrosis factor receptors (TNFRs) are a family of structurally similar membrane proteins that function as signaling pathways, activating cell death pathways or inducing gene expression involved in cellular differentiation and survival. The sword-shaped TNFR bound two different antibodies in PDB \textsl{6mhr} and \textsl{6mi2} with 60.3\% fv sequence identity. SurfBind accurately captured the differences in their CDRs and predicted the associated BCEs. A more challenging case was ADP-ribosyl cyclase, a bifunctional enzyme catalyzing an essential chemical reaction. Three SAbDab antibodies in PDB \textsl{7duo}, \textsl{3l5w}, and \textsl{8byu} bound this enzyme, and SurfBind perfectly distinguished their specific binding interfaces. Another macromolecule in humans, Interleukin 13 (IL-13), is a protein encoded by the IL-13 gene that affects immune cells in a manner similar to IL-4. Three antibodies with different binding modes to IL-13 were tested. SurfBind correctly recognized the BCE for the antibody in PDB \textsl{7rew}. However, because this antibody had over 70\% fv sequence similarity to the others, SurfBind yielded a more unified binding area for the antibodies in PDB \textsl{8blq} and \textsl{3l5w}. 

\paragraph{Explainability and Ablation Studies}
The explainability of attention~\citep{vaswani2017attention} provides a means to interpret the interactive patterns learned by SurfBind by comparing attention scores with geometric distributions in ground-truth Ab-ag complex structures. Fig.~\ref{fig:super} \textbf{c} envision for four randomly selected Ab-ag pairs, where we drew the cross-attention score maps and the distance maps between ligand and receptor residues. Higher attention scores are usually aligned with smaller minimum distances to the antibody, with a mean Spearman correlation of -0.427 between scores and distances. This shows that SurfBind captured the spatial relationship between the antibody and antigen, informing its binding decision.  

We conducted comprehensive ablation studies to assess the contributions of SurfBind components and the effects of key hyperparameters (see Fig.~\ref{fig:super} \textbf{d} and \textbf{e}). First, each technique incrementally improved the AUC-ROC on the SAbDab dataset. The SurfFormer architecture alone achieved a competitive AUC-ROC of 0.768, outperforming the DiscoTope-3.0 baseline (AUC-ROC = 0.751). Vanilla MAE-based generative pretraining improved the AUC-ROC to 0.802, and incorporating the VQ technique yielded the optimal AUC-ROC of 0.816. Moreover, the chemical property pretext task significantly increased AUC-ROC by 0.033, underscoring the necessity of physicochemical knowledge to interpret PPIs. Second, the optimal masking ratio was relatively high (50\%-70\%), as higher patch removal rates largely eliminated redundancy, making the task challenging and difficult to solve by extrapolation from visible neighboring surface patches. We also acknowledged that SurfBind was robust across downsampling ratios from 0.01 to 0.1 and benefited from a wider range of discrete latent vectors. Thirdly, if expanding the distance thresholds defining BCEs, AUC-PRs consistently escalated, \emph{e.g.}, from 0.305 to 0.578 at 10$\textup{\AA}$, attributed to AUC-PR's higher sensitivity to class imbalance than AUC-ROC. This trend reflects task relaxation rather than improved localization precision. 

\paragraph{Robustness to Binders' Conformational Changes}
Proteins undergo conformational changes coupled with ligand binding. These transitions occur across various lengths and time scales associated with functionally relevant phenomena. Here, we examined how these conformational changes affect our model's generalization on the Docking Benchmark (DB) 5.5 dataset~\citep{vreven2015updates}, which comprises high-quality protein-protein complex structures along with their unbound component forms. The dataset was categorized by interface root-mean-squared deviation (I-RMSD) between native and bound forms into rigid-body (162 cases), medium difficulty (60 cases), and difficult (35 cases) subsets. We designated DB5.5 as the test set and excluded any SAbDab duplicates with 50\% antigen sequence identity to avoid overlap as the training split. We then retrained SurfBind and examined its efficacy, making several notable findings in Fig.~\ref{fig:super} \textbf{f}. First, without pretraining, SurfBind experienced a 6.7\% overall performance decline when transferring from bound to native structures. Particularly, the loss was 15.6\% for highly flexible antigens (average I-RMSD = 3.48\textup{\AA}). Secondly, sites were easier to recognize for antigens that experienced smaller conformation changes. SurfBind attained an AUC-ROC of 0.886 for rigid-body cases, surpassing 0.793 for highly flexible ones. Thirdly, pretraining significantly improved SurfBind's robustness to unbound structures, increasing the AUC-ROC from 0.759 to 0.794. We hypothesized that the pretraining dataset contained numerous monomers, enabling SurfBind to learn surface distributions of unbound forms and reduce the transfer-learning gap.

\paragraph{Docking with Predicted Binding Sites.} To assess whether improved binding site prediction translates into practical gains in downstream modeling, we evaluate the effect of SurfBind-predicted epitopes on antibody–antigen docking accuracy. Following the DockGPT protocol~\citep{mcpartlon2023deep}, we incorporate predicted binding sites as spatial constraints during docking and compare against blind docking and ground-truth epitope guidance. As detailed in App.~\ref{app:dock}, SurfBind-guided docking consistently improves DockQ success rate from 26.1\% to 38.0\% and reduces both interface and ligand RMSD relative to blind docking, recovering a substantial fraction of the performance achieved with native epitopes. These results demonstrate that SurfBind predictions are accurate enough to meaningfully constrain docking and improve structural modeling outcomes.

\subsection{SurfBind is a Good Unsupervised Learner} 
Due to the scarcity of experimentally determined structures, PLMs have been pretrained on large unlabeled protein sequences like UniProt~\citep{lin2022language}. Despite efforts to pretrain on unlabeled 3D structures~\citep{wu2022pre}, there has been a notable lack of initiatives targeting protein molecular surfaces, which directly affect biomolecular interactions and function. SurfBind bridges this gap by introducing a generalized MAE variant tailored for molecular surfaces. Unlike conventional MAE, which uses a single embedding for masked tokens, SurfBind uses discrete variables to parameterize the representations of masked surface patches. Each masked patch's representation is replaced by the closest codebook vector in the Euclidean space before fed into the decoder. This VQ offers two key advantages: (1) Following signal processing principles, VQ achieves substantial data compression with minimal loss of geometric and physicochemical surface information. (2) Replacing individual data points with representative codebook vectors reduces noise, resulting in smoother and more robust patch-level representations. In addition to VQ, SurfBind introduces two higher-order pretext tasks beyond just coordinate reconstruction: restoring critical surface geometric properties and chemical properties computed efficiently from the compact molecular surface boundary atoms (Fig.~\ref{fig:unsper} \textbf{d}, \textbf{e}, and \textbf{f}).

\paragraph{Unsupervised Representations for BCE Discovery} 
In SurfBind, input surfaces were partitioned into patches, each mapped to a discrete codebook cluster, enabling analysis of shared patch characteristics within clusters before supervised fine-tuning. We extracted the representations of all surface patches in the SAbDab-test set and mapped them to the corresponding codebook vectors and cluster assignments. We then summarized the proportions of epitopes and non-epitopes across clusters and visualized their distributions in Fig.~\ref{fig:unsper}. As envisioned in Fig.~\ref{fig:unsper} \textbf{a}, nearly 6,000 out of 10,000 codebook clusters had a zero epitope ratio, illustrating that patches in those clusters were not prone to being BCEs. Conversely, 72 clusters had an epitope ratio exceeding 50\%, suggesting that patches in those clusters were more likely to be BCEs. Moreover, incorporating the chemical property pretext task during pretraining (refer to Fig.~\ref{fig:unsper} \textbf{b}) further improved the discrimination between BCEs and non-BCEs based purely on clusters. Moreover, as the number of codebook clusters increased from 10 to 10,000, BCEs became concentrated in a small fraction of the total number of clusters. Notably, when raising the number of cluster categories to 10,000, only about 90 patch clusters were observed for the SAbDab-test set. Under these circumstances, approximately 50 of 90 clusters lacked epitopes, and 10 clusters were enriched for BCEs. By utilizing each cluster's epitope ratio as the predicted score for all patches in that cluster, the unsupervised SurfBind achieved an AUC-ROC of 0.695 on SAbDab-test, competitive with the leading sequence-based model AF-Multimer (AUC-ROC = 0.698) and outperforming surface-based MaSIF and dMaSIF baselines. As discussed, VQMAE generalized the vanilla MAE by introducing more flexible discrete token vectors in the latent space. VQMAE reduces to MAE in the extreme case of a single codebook cluster. Augmenting the number of codebook clusters provided another perspective on VQ's effectiveness. 

\paragraph{Residue Distributions across Patch Clusters}
We analyzed the residue-type distributions across various patch clusters. For each surface patch, we associated its cluster assignment with the nearest residue. We then tallied the occurrences of each residue type within the same patch cluster. To mitigate bias arising from variations in residue-type frequencies, residue counts were normalized to the total count for each residue type across all clusters. This allowed us to derive an "averaged preference score" reflecting the residue number distribution after analyzing over 40 multi-binder antigens in Fig.~\ref{fig:unsper} \textbf{c}. We revealed that four patch clusters exhibited distinct preferences across hydropathy levels. Some clusters were enriched for hydrophobic residues, while others favored hydrophilic residues. This suggests that during pretraining, SurfBind effectively captured the physicochemical characteristics of molecular surfaces, as reflected in the residue compositions of the learned patch clusters.

\section{Conclusion} 
Protein interactions with other biomolecules are fundamental to their function in most biological processes. This study presents SurfBind, a novel surface-based structural pretraining method that extracts valuable information from large-scale collections of unlabeled molecular surfaces. We demonstrate its effectiveness across several critical and challenging downstream tasks.

\section{Limitations and Ethical Considerations}
Our approach depends on the quality of available 3D structures and may underperform for highly flexible proteins or inaccurate structural models. Predictions are probabilistic and require experimental validation. This work is intended for responsible scientific research and is not a substitute for clinical or experimental decision-making.

\bibliographystyle{ACM-Reference-Format}
\balance
\bibliography{cite}

\newpage
\appendix
\onecolumn

\section{Experimental Details}
\label{app:exp}
\subsection{Training Process}
\paragraph{Implementation Details.} We implemented all experiments on 4 H100 GPUs, each with 80G memory. During pretraining, SurfBind was trained using the Adam optimizer~\citep{kingma2014adam} with a weight decay of $5.e-3$ and $\beta_1=0.9$ and $\beta_2=0.999$. A ReduceLROnPlateau scheduler was employed to automatically adjust the learning rate with a patience of 5 epochs and a minimum learning rate of $1e-7$. The batch size was set to 32, and an initial learning rate was $1.e-4$. The maximum number of iterations was 200K, with a 10K warm-up, and the validation frequency was 1K iterations. The random seed was fixed as 2023. Moreover, we empirically calculate the overlap ratio across all patches and observe a low score of 5.34\%.

\paragraph{Dataset Details.} For the pretraining dataset, we leverage PDB-REDO that contains refined X-ray structures in the Protein Data Bank (PDB)~\citep{liu2015pdb}. Here, we followed the scheme of~\citet{luo2023rotamer} and clustered the protein chains at 50\% sequence identity, yielding 38,413 chain clusters. These clusters were randomly split into training, validation, and test sets at ratios of 95\%, 0.5\%, and 4.5\%, respectively. During the generative pretraining stage, the data loader first randomly selected a cluster, then randomly selected a chain from that cluster to ensure balanced sampling. Next, a portion (50\% - 70\%) of patches in the molecular surface of the chosen chain was randomly masked. Finally, the feature extractor was required to restore both the low-order (\emph{e.g.}, point coordinates) and high-order (\emph{e.g.}, surface geometry and physicochemical characteristics) properties of the masked surface patches. 

For antigens with multiple antibody binders, the same antigen sequence may appear across different splits only with distinct antibodies, while no antibody–antigen complex is duplicated across splits. In contrast, single-binder antigens are confined to a single split, preventing antigen-level leakage while enabling evaluation of antibody-specific epitope generalization.

\subsection{Hyperparameters}
\paragraph{Hyperparameter Search Space.} At the beginning, we adopted a random search to find the best combination of hyperparameters for the backbone architecture SurfFormer in three different downstream tasks with only supervised learning. We then fixed these hyperparameter subsets to build three backbone architectures and further explored the hyperparameters for VQMAE-style pretraining. 
\begin{table}[ht]
    \caption{Hyperparameters setup for SurfBind}
    \centering
     \resizebox{0.65\columnwidth}{!}{
    \begin{tabular}{lll}\toprule
    Hyperparameters Search Space & Symbol & Value \\ \midrule
    \textbf{Surface Generation and Partition} \\ 
    Upsampling Ratio & $\eta$ & [10, 20] \\ 
    Surface Radius & $r$ & [1.05]  \\ 
    FPS Downsampling Ratio & $\rho$ & [0.01, 0.02, 0.05, 0.1]  \\ 
    Number of Nearest Point in a Patch & $K_{\mathrm{p}}$ & [10, 25, 50, 100] \\ 
    \textbf{Backbone Achitecture} \\ 
    Dimension of Residue Type Embedding in GeoAN & $\phi_R$ &  [16] \\
    Dimension of Point Chemical Features in GeoAN & $\phi_h$ & [4, 8, 16]\\
    Dimension of Patch Chemical Features in SurfFormer & $\phi_\mathrm{p}$ & [16, 64, 128]\\ 
    Number of Nearest Residues in GeoAN & $K_{\mathrm{res}}$ & [8, 16] \\ 
    Gaussian Kernel Size & $\sigma_\mathbf{n}$ & [9\textup{\AA}, 12\textup{\AA}] \\ 
    Radius in the Filters & $\sigma_{\mathrm{d}}$ & [9\textup{\AA}, 12\textup{\AA}] \\ 
    Scaler Coefficient in Radial Basis Function & $\gamma$ & [5, 10]\\
    Number of Nearest Patches for Interpolation & $K_\textrm{in}$ & [1, 3, 5]\\
    Layer Number of Point Cloud Networks & $L_1$ & [1, 2, 3] \\
    Layer Number of SurfFormer & $L_2$ & [1, 2, 3] \\
    Number of Heads in SurfFormer & -- & [1, 4, 8] \\ 
    Dropout Rate in Point Cloud Network & -- & [0.0, 0.1] \\ 
    Dropout Rate in SurfFormer & -- & [0.0, 0.1] \\ 
    \textbf{Generative Pretraining Setup} \\
    Masking Ratio & $\delta$ & [30\%, 50\%, 60\%, 70\%, 80\%]\\
    Codebook Size & $N_B$ & [100, 1000, 10000] \\
    Target Temperature  & $\tau$ & [0.0625] \\
    Reconstruction Loss Weight & $\nu_1$ & [0.5, 1.0] \\ 
    Curvature Loss Weight & $\nu_2$ & [0.5, 1.0] \\
    Chemical Property Loss Weight  & $\nu_3$ & [0.5, 1.0]\\
    KL-divergence Loss Weight  & $\nu_4$ & [0.1, 0.5, 1.0]\\
    Coarse-to-fine Balance Term & $\zeta$  &  [0.5, 1.0]\\
    \textbf{Training Setup} \\
    Batch Size & -- & [16, 32]\\ 
    Initial Learning Rate & -- & [5e-4, 1e-4, 5e-5, 1e-6]\\ 
    Number of Warmup Iterations & -- & [5K, 10K]\\  \bottomrule 
    \end{tabular}}
    \label{tab:hyper}
\end{table}

\paragraph{Evaluation Metrics and Protocols.} To estimate the prediction performance of the benchmarked predictors, we used a variety of well-established performance metrics, including the balanced accuracy (BAcc), the F-score, the area under the receiver operating characteristic curve (AUC-ROC), and the area under the precision-recall curve (AUC-PR). 

Consistent with the findings of~\citet {cia2023critical}, our investigation revealed that as the threshold for identifying surface residues increased, all methods not based on surface assessment yielded consistently lower scores. This suggests that, as we narrow our evaluation to surface residues only, these methods perform worse. Conversely, including buried residues in the surface category simplifies predictions by artificially amplifying the disparity between epitopes and non-epitopes. This enrichment primarily arises from a greater abundance of hydrophobic residues among non-epitopes, relative to epitopes, as delineated by an additional RSA threshold.

\subsection{Extended Introduction of SurfFormer++}
\label{app:surfformer}
\begin{figure*}[t]
    \centering
    \includegraphics[width=\textwidth]{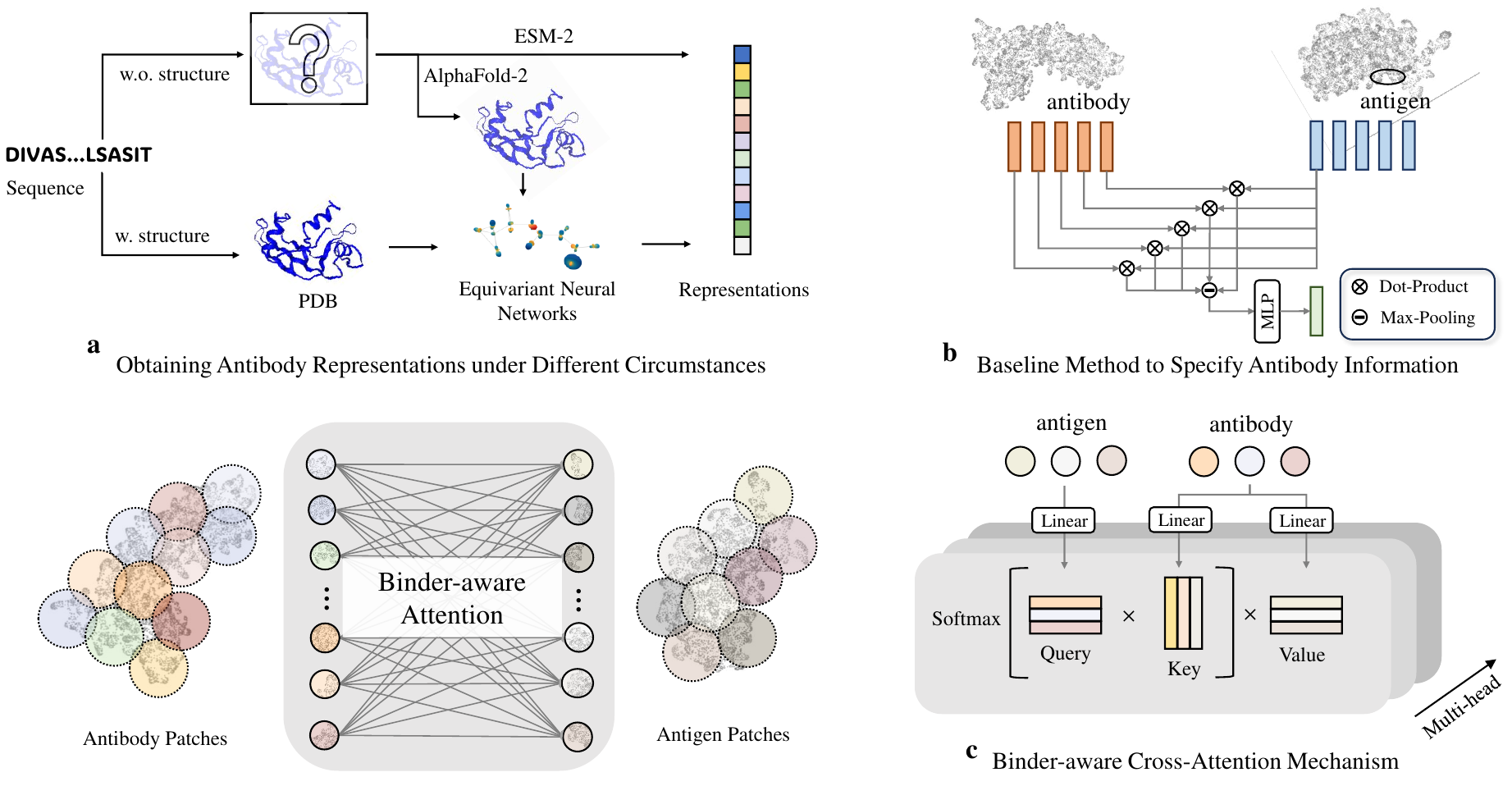}
    \caption{\textbf{a}. Different approaches to acquiring antibody representations when antibody structures may be inaccessible.  \textbf{b}. A simple and general scheme to enable antibody-specificity. \textbf{c}. The cross-attention method incorporates the antibody information into antigens. } 
    \label{fig:app_model}
\end{figure*}
We highlighted that SurfFormer++ employs cross-attention to connect the antibody and the antigen and achieve binder-awareness (see Fig.~\ref{fig:app_model} \textbf{c}). Moreover, our architecture applies to a wide range of scenarios, even when antibody structures are unavailable. In that situation, we can rely on PLMs to extract the protein-level representation of the antibody and forward it to the cross-attention calculation as the key and query (see Fig.~\ref{fig:app_model} \textbf{a}). Additionally, we proposed a simple technique to make traditional BCE-based prediction models partner-specific. Specifically, we used a dot product and a max-pooling operation to exchange information between the ligand and the receptor (see Fig.~\ref{fig:app_model} \textbf{b}).

\section{Benchmarking Baselines}

Recent years have seen exponential growth in BCE data, prompting rapid advances in machine learning (ML) methods for predicting ADs. They use the physicochemical properties of amino acids as descriptors to rapidly and efficiently identify potential epitopes as vaccine candidates, thereby reducing the burden of the BCE mapping process by narrowing the list of candidate epitopes for experimental trials. 
Preliminary endeavors use empirically computed energy terms and contact-frequency-based features as direct inputs to ML models such as support vector machines (SVMs)~\citep{el2008predicting} and random forests (RFs)~\citep{bukhari2021using}, but achieve limited accuracy due to the lack of 3D complementary information. Meanwhile, some methods rely on sequence conservation or residue coevolution but often perform poorly for shallow sequence alignments~\citep{shashkova2022sema}. Approaches such as AF-Multimer, which are centered on de novo complex folding, concurrently reveal interfaces and subunit conformations. However, they are limited to PPIs, are slower than structure-based interface prediction, and can fail if the folding protocol falters.

To address this constraint, geometric DL, an umbrella term that generalizes networks to Euclidean or non-Euclidean domains, has emerged as a promising avenue for modeling macromolecular structures. Adapting it to protein structures necessitates defining an appropriate protein representation. Early studies~\citep{wang2020protein} integrate detailed atomic spatial information by isolating a 3D voxel grid around the interface region and employing convolutional neural networks (CNNs). Subsequent works~\citep{da2022epitope3d} describe antigens as 3D graphs, treating residues as vertices and the distance angles between them as edges. This condenses complex 3D information into compact signatures while preserving binding-related spatial features, and the signatures are processed by graph neural networks (GNNs) with E(3) or SE(3) equivariance and symmetry. Embracing the premise that every surface residue may be immunogenic, a prevailing line of research focuses on molecular surfaces. MaSIF~\citep{gainza2020deciphering} pioneers the use of meshes, defining the patch as a region on a solvent-excluded protein surface with a fixed geodesic radius around a potential contact point to predict interactions. dMaSIF~\citep{sverrisson2021fast} streamlines the process by sampling atomic point clouds, alleviating the need for pre-calculations.

\subsection{Sequence-based Algorithms}
\textbf{AF-Multimer} predicts protein complexes from MSAs with impressive performance in general protein–protein docking tasks. In our experiments, we used the AF-Multimer implementation provided by \texttt{ColabFold}, which performs an MSA search using \texttt{MMseqs2}. Following the default ColabFold protocol, we enabled MSAs during inference. Since paired MSAs are generally unavailable for antibody–antigen systems, we used \emph{unpaired} MSAs for the two chains, consistent with common benchmarking practice. We used the default parameters: 10 recycles and 5 predicted models. For downstream evaluation, we extracted the putative interfaces of the AF-Multimer predictions (\emph{i.e.}, residue–residue contacts within 4\textup{\AA}) and averaged the results across the 5 models. The code used in this paper was obtained from~\url{https://github.com/sokrypton/ColabFold}. 
\textbf{CBTOPE}~\citep{ansari2010identification} is the first attempt to predict conformational BCEs from an amino acid sequence. It trained support vector machine (SVM) models, and we used the web server at~\url{http://www.imtech.res.in/raghava/cbtope/}. 
\textbf{Seppa-3.0}~\citep{zhou2019seppa} used a logistic regression model to present a raw antigenicity score for each surface residue based on the micro-environment features, such as glycosylation triangles and glycosylation-related amino acid indexes. This score was then calibrated by the overall tendency of neighboring residues. We accessed its program at~\url{http://www.badd-cao.net/seppa3/}. 
\textbf{BepiPred-2.0}~\citep{jespersen2017bepipred} replied on a random forest algorithm to forecast BCEs from antigen sequences. It analyzed the residues using hydrophobicity and polarity measurements, along with their volume, RSA, and predicted secondary structure. We leveraged its web server at~\url{https://services.healthtech.dtu.dk/services/BepiPred-2.0/} for validation. 
\textbf{BepiPred-3.0}~\citep{clifford2022bepipred} is a sequence-based tool that uses numerical representations from the protein language model ESM-2~\citep{lin2022language} to significantly improve the accuracy of both linear and conformational BCE prediction. We used its freely available web server and a standalone package at~\url{https://services.healthtech.dtu.dk/services/BepiPred-3.0/} to navigate the results. 

\subsection{Structure-based Algorithms}

\textbf{Epi-EPMP}~\citep{del2021neural} leverages a message-passing network to perform joint paratope-epitope prediction. However, its code is not publicly available. 
\textbf{ScanNet}~\citep{tubiana2022scannet} is an end-to-end, interpretable geometric DL model that learned features directly from 3D structures. It builds representations of atoms and amino acids based on the spatio-chemical arrangement of their neighbors. ScanNet is a versatile, powerful, and interpretable model suitable for functional site prediction tasks, and we used its publicly available web server at~\url{http://bioinfo3d.cs.tau.ac.il/ScanNet/index_real.html}. 
\textbf{Epitope3D}~\citep{da2022epitope3d} used the concept of graph-based signatures to model epitope and non-epitope regions as graphs and extracted distance patterns that were used as evidence to train and test predictive models. We submitted jobs via an API at~\url{https://biosig.lab.uq.edu.au/epitope3d/api} for validation. 
\textbf{PeSTo}~\citep{krapp2023pesto} is a geometric Transformer that acts directly on atomic coordinates labeled only with element names. Its low computational cost enabled processing of large volumes of structural data, and we used its publicly available code at ~ \ url {https://github.com/LBM-EPFL/PeSTo}.
\textbf{SEMA-2.0}~\citep{shashkova2022sema} fine-tuned ESM-1v~\citep{rives2021biological} and ESM-IF1~\citep{hsu2022learning} models to predict residues comprising BCEs by providing an interpretable score corresponding to the expected number of contacts of an amino acid residue with the target antibody. Two models were independently fine-tuned, yielding SEMA-1D and SEMA-3D. The author reported better performance of SEMA-3D, which we validated on the web server at~\url{https://sema.airi.net/prediction_analysis}.
\textbf{DiscoTope-3.0}~\citep{hoie2024discotope} is a structure-based BCE prediction tool exploiting inverse folding representations generated from either AlphaFold predicted or solved structures. It adopted the XGBoost architecture, which was trained on both predicted and solved antigen structures using a positive-unlabelled learning ensemble approach. This enabled large-scale prediction of epitopes even when solved structures were unavailable. We used its web server at~\url{https://services.healthtech.dtu.dk/services/DiscoTope-3.0/}. 
\textbf{Pair-EGRET}~\citep{alam2023pair} is an edge-aggregated graph attention network (GAT) that leveraged the features extracted from pretrained Transformer-like models to accurately predict pairwise PPI sites. It used a k-nearest-neighbor graph to represent the three-dimensional structure of a protein and employed cross-attention on top of a Siamese network to accurately identify interface residues between protein pairs. We re-run its code at~\url{https://github.com/1705004/Pair-EGRET}. 
\textbf{WALLE}~\citep{liu2024asep} models each antibody–antigen complex as two residue-level graphs (antibody and antigen), with nodes initialized by concatenating ESM embeddings and structural features. Each graph is encoded independently using stacked GNN layers, after which antibody–antigen residue pairs are combined in a cross-graph decoder to predict inter-residue contacts. We reproduce their method following the repository at~\url{https://github.com/biochunan/AsEP-dataset}. 

\subsection{Surface-based Algorithms}
\textbf{MaSIF-Site}~\citep{gainza2020deciphering} is a conceptual framework that uses a geometric DL method to capture fingerprints important for specific biomolecular interactions. It took as input mesh-based representations of a protein surface and relied on hand-crafted chemical and geometric features, which must also be pre-computed and stored on the hard drive. We used the recommended parameters for data processing: Circular patches of 12\textup{\AA} radius were computed from the surfaces of interacting proteins using the MaSIF data preparation module. The patch data structure was a grid of 80 bins, with 5 angular and 16 radial coordinates. We leveraged the code at~\url{https://github.com/LPDI-EPFL/masif}.
\textbf{dMaSIF-Search}~\citep{sverrisson2021fast} extended MaSIF by bypassing the pre-computation of physicochemical features and instead calculating molecular surfaces directly from the atomic point cloud in real-time. The model's input had a data structure similar to that of a 12\textup{\AA} patch, in which each surface point was represented as a one-hot encoding of surface chemicals, together with Gaussian and mean curvatures. We retained the code from~\url{https://github.com/FreyrS/dMaSIF.git} and trained from scratch for evaluation. 
\textbf{SEPPA-mAb}~\citep{qiu2023seppa} appended a fingerprints-based patch model to {Seppa-3.0}~\citep{zhou2019seppa}, considering the structural and physicochemical complementarity between a possible epitope patch and the CDRs of monoclonal antibodies. We leveraged its web server at~\url{http://www.badd-cao.net/seppa-mab/}. 

\subsection{Limitations and Future Work}
To the best of our knowledge, the developed SurfBind is the first attempt to conduct generative pretraining purely on molecular surfaces. It outperforms state-of-the-art empirical and ML-based protein-scoring functions in identifying antibody-specific viable BCEs. In spite of its promising progress, there is still some space left for future explorations. 
First, more abundant databases can be exploited in our framework. The powerful structure prediction methods~\citep{varadi2022alphafold}
Alphafold-Database~\citep{varadi2024alphafold} 
Second, PLMs have shown effectiveness in many protein-related tasks, It might be beneficial if both the PLM and the geometric encoder are tuned. 
Second, our protein graphs are built in a residue-level manner. However,~\citep{jin2022antibody} has already demonstrated that atom-level modeling significantly surpasses the residue-level one. Therefore, it is undeniable that the performance of our HTP will be improved dramatically if an atom-level protein graph is considered.

\begin{figure*}[t]
    \centering
    \includegraphics[width=\textwidth]{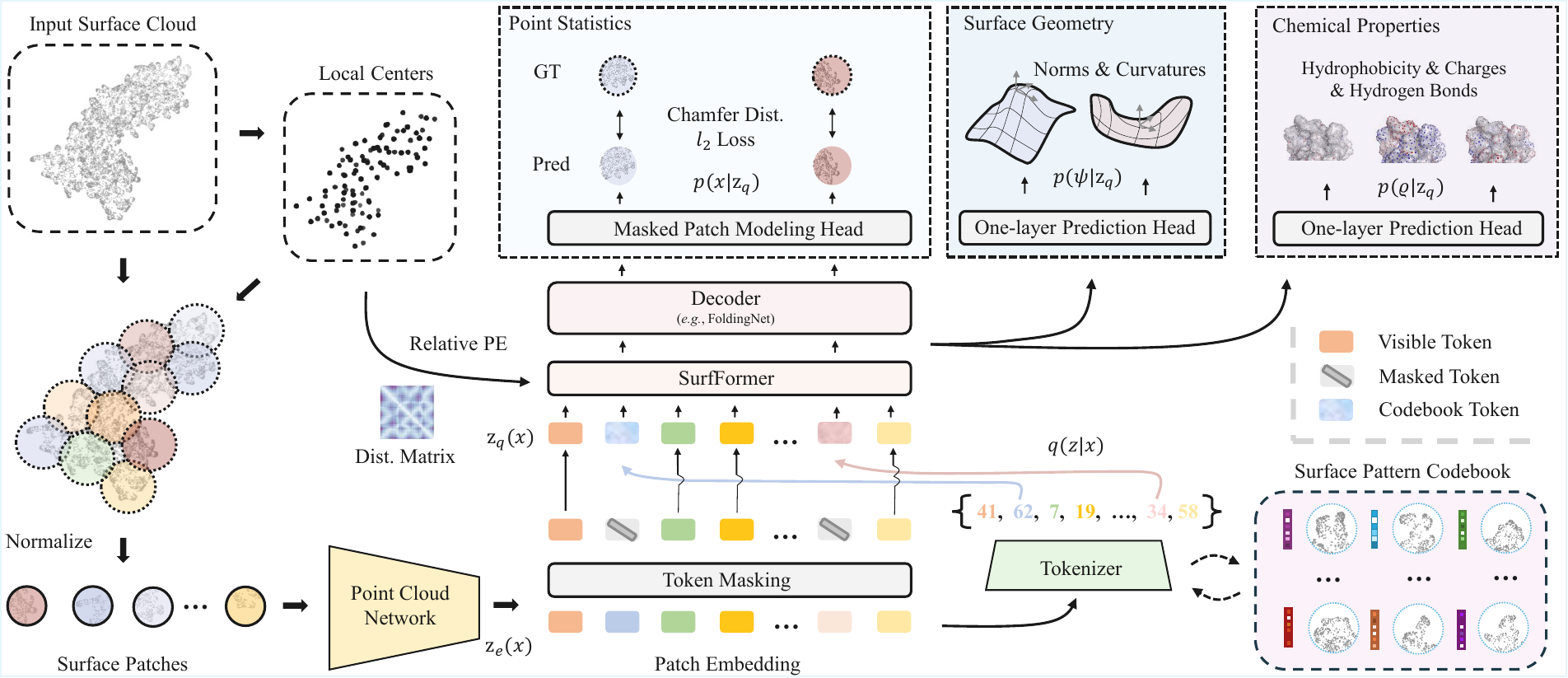}
    \caption{\textbf{The overall pipeline of our unsupervised SurfBind method.} The input surface point cloud is first preprocessed into ordered patches using the farthest point search (FPS) algorithm and Morton codes. Patch-level representations are then extracted by a point cloud network. A portion of patches is randomly masked, and their features are replaced with vectors from a relaxed codebook. Both visible patch embeddings and sampled codebook vectors are forwarded to SurfFormer to gain a global surface understanding. Finally, three pretext tasks are proposed as pretraining objectives: reconstructing the coordinates of masked center points, predicting local surface geometry, and forecasting critical surface chemical properties.}
    \label{fig:model_vae}
\end{figure*}

\section{Docking Performance with SurfBind-Guided Binding Sites}
\label{app:dock}
To evaluate whether improved binding site prediction translates into gains in downstream structural modeling, we assess the impact of SurfBind-predicted epitopes on protein–protein docking accuracy. Docking provides a practical and stringent test of binding site quality, as accurate site localization can substantially reduce the conformational search space and improve pose selection.

Following the experimental protocol of DockGPT~\citep{mcpartlon2023deep}, we incorporate SurfBind-predicted epitopes as spatial constraints within the docking pipeline. We compare three settings: (i) blind docking without binding site information, (ii) docking guided by SurfBind-predicted epitopes, and (iii) docking guided by ground-truth epitopes. Docking quality is evaluated using DockQ~\citep{basu2016dockq}, along with interface RMSD (I-RMSD) and ligand RMSD (L-RMSD),  with results reported per target by taking the best prediction among the top-ranked poses. 

Specifically, we assessed docking performance on two standard benchmarks: the Antibody Benchmark (Ab-Bench), comprising 46 antibody–antigen complexes with unbound structures, and a held-out subset of DB5.5, containing 42 non-redundant protein–protein complexes that are sequence-disjoint from the training data. To account for stochasticity when binding-site constraints were provided, each target was evaluated over multiple independent runs with different random interface samplings, and metrics were averaged per complex. 

Results are summarized in Table~\ref{tab:docking_results}. Incorporating SurfBind-predicted epitopes yields consistent improvements over blind docking across all metrics. In particular, the docking success rate increases from 26.1\% under blind docking to 38.0\% when guided by SurfBind predictions. Consequently, both I-RMSD and L-RMSD percentiles decrease, indicating more accurate interface reconstruction and ligand placement. While docking guided by ground-truth epitopes achieves a higher performance (54.3\% success rate), SurfBind recovers a substantial fraction of this gain without access to native binding site annotations.

These results indicate that SurfBind predictions are sufficiently accurate to serve as effective docking constraints, leading to measurable improvements in docking accuracy. Beyond standard epitope prediction benchmarks such as SAbDab and DB5.5, this experiment demonstrates that SurfBind provides practical benefits when integrated into downstream antibody–antigen docking pipelines, supporting its utility for structure-based interaction modeling. 
While we do not perform formal hypothesis testing, the performance gains are robust across complexes and are not driven by a small subset of outliers, indicating consistent improvement across targets rather than isolated cases.
\begin{table}[h]
\centering
\caption{Comparison of antibody--antigen docking performance following the \emph{DockGPT} protocol. 
Results are reported in terms of DockQ success rate (SR, higher is better) [B] and I-RMSD/L-RMSD percentiles (lower is better). 
The top block reports \textbf{blind docking} (no epitope information provided). 
The bottom block reports \textbf{epitope-guided docking}, where DockGPT is supplied either with SurfBind-predicted epitopes or ground-truth epitopes.}
\label{tab:docking_results}
\begin{tabular}{l c ccc ccc}
\toprule
& \textbf{SR} $\uparrow$ & \multicolumn{3}{c}{\textbf{I-RMSD} $\downarrow$} & \multicolumn{3}{c}{\textbf{L-RMSD} $\downarrow$} \\
\cmidrule(lr){3-5} \cmidrule(lr){6-8}
& & 25$^{\text{th}}$ & 50$^{\text{th}}$ & 75$^{\text{th}}$ & 25$^{\text{th}}$ & 50$^{\text{th}}$ & 75$^{\text{th}}$ \\
\midrule
\multicolumn{8}{l}{\textbf{Blind Docking}} \\
EquiDock       & 0.0\%   & 11.6 & 13.7 & 16.8 & 31.9 & 41.1 & 51.0 \\
ZDock          & 2.2\%   & 10.1 & 12.8 & 17.0 & 23.9 & 28.2 & 39.3 \\
PatchDock      & 0.0\%   & 12.0 & 13.9 & 15.5 & 26.1 & 32.2 & 46.3 \\
HDock          & 2.2\%   & 12.5 & 15.6 & 19.8 & 24.0 & 47.3 & 58.5 \\
AF-Multimer    & 28.3\%  & 1.9  & 9.3  & 14.7 & 12.2 & 22.6 & 36.0 \\
DockGPT        & 26.1\%  & 2.5  & 9.2  & 12.1 & 8.2  & 19.5 & 25.4 \\
\midrule 
\multicolumn{8}{l}{\textbf{Epitope-Guided Docking}} \\
DockGPT + SurfBind & 38.0\% & 2.1 & 7.3 & 9.7  & 6.4 & 14.9 & 22.6 \\
DockGPT + ground-truth & 54.3\% & 1.6 & 3.1 & 6.8  & 4.6 & 9.5  & 20.6 \\
\bottomrule
\end{tabular}
\vspace{-1em}
\end{table}

\end{document}